\documentclass[10pt, letterpaper, twocolumn]{article}

\usepackage[T1]{fontenc}
\usepackage{truthtensor_whitepaper}
\usepackage[style=ieee]{biblatex}

\title{TruthTensor: Evaluating LLMs through Human Imitation on Prediction Market under Drift and Holistic Reasoning} 

\author{\vspace{1em}
  {\normalsize Shirin Shahabi,} {\normalsize Spencer Graham, and} {\normalsize Haruna Isah}\\
  {\normalsize [\href{mailto:shirin@inferencelabs.com}{shirin,} \href{mailto:spencer@inferencelabs.com}{spencer, and}  \href{mailto:haruna@inferencelabs.com}{haruna}]}@inferencelabs.com\\
  {\normalsize Inference Labs Inc.} 
}

\date{\today}

\begin{document}

\twocolumn[
\begin{@twocolumnfalse}
\maketitle
\begin{abstract}
\noindent
Evaluating language models and AI agents remains fundamentally challenging because static benchmarks fail to capture real-world uncertainty, distribution shift, and the gap between isolated task accuracy and human-aligned decision-making under evolving conditions. This paper introduces TruthTensor, a novel, reproducible evaluation paradigm that measures reasoning models not only as prediction engines but as human-imitation systems operating in socially-grounded, high-entropy environments. Building on forward-looking, contamination-free tasks, our framework anchors evaluation to live prediction markets and combines probabilistic scoring to provide a holistic view of model behavior. TruthTensor complements traditional correctness metrics with drift-centric diagnostics and explicit robustness checks for reproducibility. It specify human vs. automated evaluation roles, annotation protocols, and statistical testing procedures to ensure interpretability and replicability of results. In experiments across 500+ real markets (political, economic, cultural, technological), TruthTensor demonstrates that models with similar forecast accuracy can diverge markedly in calibration, drift, and risk-sensitivity, underscoring the need to evaluate models along multiple axes (accuracy, calibration, narrative stability, cost, and resource efficiency). TruthTensor therefore operationalizes modern evaluation best practices, clear hypothesis framing, careful metric selection, transparent compute/cost reporting, human-in-the-loop validation, and open, versioned evaluation contracts, to produce defensible assessments of LLMs in real-world decision contexts. We publicly released TruthTensor at \href{https://truthtensor.com/}{https://truthtensor.com}. 
\end{abstract}
\tableofcontents
\vspace{1em}

\end{@twocolumnfalse}
]

\section{Introduction}

Artificial intelligence (AI) evaluation has traditionally been anchored in static metrics and benchmark datasets designed to assess models' performance in predefined tasks, such as general problem-solving \cite{chollet2019measure}, code generation \cite{chen2021evaluating}, or mathematical reasoning \cite{cobbe2021training}. The fundamental issue with these traditional benchmarks is their reliance on closed-world evaluations, where AI models are tested on a fixed set of tasks or datasets that often contain historical information or established patterns. In these settings, models can perform exceptionally well by memorizing data or applying pattern-matching techniques, but these metrics fail to capture a model's ability to reason, adapt, or learn in novel contexts. As AI systems grow more complex and exhibit behaviors that seem to mimic human cognition, such as abstract reasoning, dynamic problem-solving, and real-time interaction, the static nature of these benchmarks becomes increasingly inadequate for assessing the true breadth and depth of AI capabilities.

In recent years, the AI community has acknowledged the limitations of these traditional evaluation frameworks. The reliance on outdated benchmarks for specialized tasks, while still necessary for certain applications, no longer suffices for evaluating the full range of cognitive abilities that next-generation models exhibit \cite{zhang2025llmeval}. This gap is highlighted by initiatives like Chatbot arena\cite{chiang2024chatbot}, CORE-Bench \cite{siegel2024core}, Futurex \cite{zeng2025futurex} and Humanity's Last Exam \cite{phan2025humanity}, which attempts to assess models' capabilities in more comprehensive and real-world contexts. However, even such ambitious projects are often limited by the underlying problem that they still measure how well models parse or reproduce knowledge derived from historical datasets, rather than how they might reason, extrapolate, or adapt to new, unseen scenarios. 

\subsection{TruthTensor: A Paradigm Shift from Prediction to Human Imitation}
In response to the limitations of traditional AI evaluation frameworks, this work introduces TruthTensor, a novel approach that fundamentally reframes the evaluation question. Rather than asking “How well can this model predict future events?”, a question that conflates memorization, pattern matching, and genuine reasoning, TruthTensor asks: “How well does this model imitate human reasoning, calibration, and narrative coherence when confronted with evolving, uncertain, socially-grounded scenarios?”. This shift from prediction evaluation to human imitation assessment represents a fundamental departure from static evaluation methods that test information retrieval from historical datasets.

TruthTensor integrates LLMs with streaming real-time prediction market data, challenging their ability to reason through complex, high-entropy environments where ground truth has yet to be established. The framework measures not only whether models produce accurate forecasts, but how their reasoning processes, confidence calibration, and narrative stability compare to human market participants. Anchoring evaluation to prediction markets which aggregate financially backed, continuously updated crowd expectations, provides TruthTensor with a dynamic, socially grounded reference point for assessing LLMs human imitation capabilities.

\subsection{Key Differentiators}
TruthTensor differs from existing benchmarks in several critical ways:
\begin{itemize}[noitemsep, topsep=0pt]
\item Human Imitation Focus: Evaluation centers on how well LLMs replicate human reasoning patterns, calibration, and narrative coherence, not merely prediction accuracy.
\item Drift-Centric Design: The framework places primary emphasis on measuring narrative drift, temporal inconsistency, and reasoning confidence decay, dimensions largely ignored by existing benchmarks.
\item Contamination-Free Construction: Evaluates only forward-looking events, thereby eliminating data contamination by construction, a fundamental weakness of static benchmarks.
\item Baseline Independence: Baseline models provide reference points independent of rolling-window calibration, ensuring fair comparison across models with different training histories.
\item Instruction Locking: Prompt specifications are versioned and locked, ensuring reproducibility and preventing prompt engineering from masking model limitations.
\item Holistic Evaluation: Metrics span correctness, risk assessment, temporal coherence, calibration, and drift magnitude, providing a comprehensive view of model capabilities.
\item Specified Evaluation Categories: Events are categorized by risk profile, domain, temporal horizon, and market liquidity, enabling targeted analysis of model strengths and weaknesses.
\end{itemize}
Together, these design choices motivate the background concepts and methodological foundations introduced in the following section.

\section{Background}
\label{sec:background}
LLM evaluation extends far beyond assigning scores on fixed benchmarks. Evaluation serves multiple purposes including comparison, diagnosis, selection, and deployment assurance. Its design is inherently shaped by who is evaluating (model builders, researchers, or end users) and why it is carried out. The Evaluation Guidebook \cite{fourrier2025evalguidebook}, highlights that no single benchmark or metric can fully characterize model quality as evaluations encode assumptions about tasks, data, prompts, and metrics, each introducing trade-offs and biases. The guidebook emphasizes the limitations of static leaderboards, the risks of overfitting to benchmarks, and the growing importance of custom, task-specific, and human-aligned evaluations. It further distinguishes between offline and online evaluation, automatic and human-in-the-loop methods, and capability versus behavior testing, while underscoring challenges such as data contamination, prompt sensitivity, and distributional shift. The guidebook motivates a more principled, transparent, and purpose-driven approach to evaluation, one that recognizes evaluation as an evolving system rather than a fixed score, providing the conceptual backdrop for alternative frameworks that prioritize robustness, interpretability, and real-world relevance.

\subsection{Reasoning Beyond Training Data: Avoiding Contamination}

Static evaluation on held‐out test sets has been shown to overstate model performance. As training corpora grow ever larger, it becomes difficult to ensure that no part of a static benchmark has leaked into a model’s training data. Recent analyses find that evaluating LLMs on fixed benchmarks is vulnerable to data contamination and leaderboard overfitting, and that traditional test sets are inevitably affected by possible data contamination \cite{zhang2025llmeval}. Forecasting-based evaluation avoids this pitfall by construction: models must predict future events that by definition did not exist during training \cite{paleka2024consistency}.

TruthTensor extends this principle by exclusively evaluating forward-looking events whose outcomes are unknown at prediction time. This design choice ensures that no model can leverage memorized information about test set outcomes, forcing evaluation to focus on genuine reasoning capabilities rather than training data recall. The framework further mitigates contamination risk through instruction locking: prompt templates are versioned and immutable, preventing researchers from inadvertently introducing test set information through prompt engineering.

\subsection{LLMs as Oracles: The Human Imitation Perspective}
A growing body of research explores the use of LLMs as Oracles \cite{oracle2025llm}, systems that provide probabilistic assessments or judgments about uncertain events. This perspective reframes LLMs not as deterministic answer generators, but as probabilistic reasoning systems that, like human experts, must navigate uncertainty, update beliefs, and express confidence appropriately. TruthTensor adopts this oracle perspective, evaluating LLMs’ ability to function as human-like probabilistic reasoners in socially-grounded contexts. 

The oracle framework is particularly relevant for prediction markets, where human participants continuously update their probability estimates based on new information, market dynamics, and narrative shifts. Through the comparison of LLM outputs to market-implied probabilities which represent aggregated human expectations, TruthTensor measures how well models replicate human-like reasoning patterns, calibration, and narrative coherence. This comparison reveals not just prediction accuracy, but the quality of the underlying reasoning process: models that exhibit human-like drift patterns, appropriate confidence calibration, and narrative stability are better oracles than those that produce static, overconfident, or narratively inconsistent outputs.

\subsection{LLM Evaluation and Benchmarks}
New frameworks for evaluating large language models (LLMs) are moving away from one-size-fits-all benchmarks toward a combination of task-specific evaluations, human-centric assessments, and robustness tests \cite{raschka2023}. Such frameworks consider the variety of ways that AI models are integrated into the real world, from their ability to understand complex instructions and engage in nuanced dialogue, to their capacity for handling ambiguous inputs and generating creative outputs.

The longstanding use of benchmarks like HumanEval \cite{chen2021evaluating}\cite{zhang2024humaneval} for coding tasks, GSM8K \cite{cobbe2021training} for mathematical reasoning, and ARC \cite{chollet2019measure} for general knowledge questions has been instrumental in providing an initial gauge for model performance. These tools, while useful for tracking progress over time, have increasingly been shown to offer a limited and outdated measure of the true cognitive capabilities of modern AI systems. Emerging discussions around AI evaluation have called for a shift toward open-world evaluation, where models are assessed on their ability to generalize, handle uncertainty, and demonstrate adaptive reasoning in ever-changing environments \cite{damani2025beyond}. This approach aims to move beyond rote knowledge retrieval and instead measure how AI systems engage with novel problem-solving scenarios that require creativity, abstraction, and domain-independent thinking. 

Recent research underscores the need for a more diverse and multi-dimensional approach to AI evaluation. For instance, the Chatbot Arena framework \cite{chiang2024chatbot} addresses the limitations of conventional single-turn evaluation of conversational AI by introducing a multi-turn, interactive benchmarking environment. In this setup, multiple conversational agents, or agents interacting with human interlocutors, engage in structured dialogues across diverse scenarios. The resulting interactions are assessed along metrics such as coherence, consistency, helpfulness, and overall conversational quality.

Humanity’s Last Exam \cite{phan2025humanity} proposes a comprehensive evaluation framework that challenges large language models (LLMs) with a broad suite of tasks designed to assess general intelligence across multiple cognitive domains. Rather than focusing narrowly on individual capabilities, such as code generation or math reasoning, the benchmark includes diverse tasks spanning logic, commonsense reasoning, real-world knowledge, and abstract problem-solving. The design of Humanity’s Last Exam emphasizes generalization and adaptability: models are evaluated on their ability to apply reasoning to new, previously unseen tasks, thereby reducing the likelihood that success stems merely from memorizing patterns in training data or overfitting to narrow benchmarks.

Futurex \cite{zeng2025futurex} introduces a dynamic forecasting benchmark designed to evaluate AI models’ ability to reason and make probabilistic predictions about future events under uncertainty. Unlike traditional static datasets, all tasks in Futurex are forward-looking, with outcomes unknown at the time of prediction, effectively preventing data contamination and memorization. The benchmark spans geopolitical, economic, and social events, requiring models to produce probabilistic forecasts evaluated via proper scoring rules such as the Brier score and log-likelihood.

ForecastBench \cite{karger2024forecastbench} is another dynamic benchmark designed to evaluate AI models’ forecasting capabilities on real-world events whose outcomes are unknown at the time of prediction. Unlike traditional static benchmarks that rely on historical data and fixed test sets, ForecastBench constructs tasks from events such as elections, economic indicators, geopolitical developments, and other measurable phenomena, ensuring that the “test set” is effectively the future. Models are required to produce probabilistic forecasts, and evaluation is based on proper scoring rules that reward calibration and reliability, such as the Brier score or log-likelihood metrics.

CORE-Bench \cite{siegel2024core} is a benchmark designed to evaluate AI agents' ability to ensure computational reproducibility in scientific research. Unlike traditional benchmarks that focus on standard tasks, CORE-Bench constructs challenges based on real-world scientific research, requiring agents to replicate experimental processes and outcomes from published papers. Evaluation is based on criteria such as code execution, data handling, and result verification, ensuring that the reproduced results match the original findings. CORE-Bench aims to improve the credibility and transparency of scientific research.

LiveCodeBench \cite{jain2024livecodebench} is a benchmark designed to evaluate the capabilities of large language models (LLMs) in code-related tasks, with an emphasis on holistic and contamination-free evaluation. Unlike traditional code-based benchmarks that often rely on static datasets or tasks, LiveCodeBench constructs diverse challenges that test LLMs on real-world programming scenarios, such as bug detection, code completion, and refactoring. The benchmark ensures that the test set is free from contamination by pre-existing model outputs, providing a more reliable assessment of LLM performance. Evaluation is based on a range of metrics, including accuracy, code efficiency, and robustness, to assess models' generalization across various programming domains.

InvestorBench \cite{li2025investorbench} is a benchmark framework proposed for evaluating AI agents’ capabilities to forecast and reason about financial markets and investment decisions. It frames evaluation as a forecasting task where models must predict asset price movements, financial outcomes, or market trends over defined horizons. Unlike static datasets, InvestorBench’s tasks are forward-looking: at prediction time, the ground truth has not yet been realized. Performance is evaluated according to probabilistic forecasts, calibration, and decision quality, capturing not just whether a model can guess the right outcome, but how well it can reason under financial uncertainty, assess risk, and update predictions over time. This benchmark explicitly targets financial‑market reasoning, bridging the gap between traditional AI evaluation (e.g., language, reasoning, classification) and real-world decision tasks where stakes, uncertainty, and temporal dynamics matter.

MIRAI \cite{ye2024mirai} is an evaluation framework (and associated environment) aimed at assessing LLMs in interactive, agentic, and decision-making contexts. Rather than evaluating models through static prompts and single-run outputs, MIRAI emphasizes agentic execution, allowing LLM-based agents to interact with external tools or environments, perform multi-step reasoning or actions, and adapt over time. The framework supports modular agent composition, enabling evaluation of not just the base language model, but the entire decision-making stack including tool usage, planning, and dynamic context handling.

Prophet Arena \cite{yang2025llm} is a dynamic benchmarking platform designed to evaluate AI agents’ forecasting capabilities in real-world, uncertain environments. Unlike static evaluation datasets, Prophet Arena emphasizes forward-looking tasks where outcomes are not yet known at the time of prediction, ensuring that models are tested on their ability to reason under uncertainty rather than memorize historical data. The framework aggregates probabilistic predictions from AI agents across a variety of socially and economically relevant events, including political, cultural, and financial phenomena. Evaluation focuses on calibration, accuracy, and adaptability over time, enabling the assessment of both short-term and long-term forecasting performance.

Agent Market Arena (AMA)\cite{qian2025agents} proposes a market-based, multi-agent evaluation ecosystem where autonomous agents interact, negotiate, or trade in simulated or real markets. Under AMA, agents are evaluated not on static benchmark performance but on their strategic behavior, adaptability, and decision-making in a shared environment with other agents. The evaluation includes how agents respond to other agents' behavior, market dynamics, and evolving information, capturing strategic reasoning, temporal adaptation, and emergent group dynamics. AMA represents a shift from static, isolated evaluation toward interactive, socially grounded assessment, and provides a valuable perspective on how AI agents might behave in settings where their decisions influence and respond to others.

LiveTradeBench \cite{yu2025livetradebench} is a benchmarking methodology designed to evaluate AI agents’ performance in live trading environments, using real or simulated financial markets. Instead of evaluating agents on offline historical data or synthetic tasks, LiveTradeBench emphasizes real-time decision‑making under uncertainty, requiring agents to consume live market data, make predictions or trading decisions, and manage risk dynamically. Evaluation metrics include not only forecast calibration and prediction quality, but also trade outcome, profit-and-loss (PnL), drawdowns, and risk-adjusted returns. This benchmark assesses an agent’s ability to reason under uncertainty, update forecasts with fresh information, and make economically rational decisions in volatile, high-stakes environments, attributes that static benchmarks fail to capture. LiveTradeBench underlines the importance of temporal coherence, risk sensitivity, and decision‑making stability in AI agents that operate in real-world financial domains.

Building on these prior efforts, TruthTensor integrates live prediction-market data to create a dynamic, continuously evolving evaluation environment. Unlike existing benchmarks, TruthTensor emphasizes real-time probabilistic reasoning and agentic decision-making, moving beyond recall or static prediction to measure how well AI models think, adapt, and forecast in high-entropy, socially relevant contexts.

As AI continues to evolve, it is clear that the field must leave behind the narrow scope of traditional metrics in favor of dynamic, flexible evaluation methodologies that can capture the true cognitive potential of these systems. The evaluation process must reflect the complexities of real-world interactions and decision-making, ensuring that AI systems are not only competent in completing tasks but are also equipped to handle the unpredictable nature of human interaction, novel scenarios, and evolving challenges.

To contextualize the contributions of TruthTensor relative to existing AI evaluation frameworks, Table \ref{tab:evaluation_frameworks} summarizes the key features, evaluation types, focus areas, and limitations of several representative benchmarks. As shown, while prior benchmarks such as Humanity's Last Exam, ForecastBench, Futurex, and Chatbot Arena provide valuable insights into general intelligence, probabilistic forecasting, or multi-turn conversational performance, TruthTensor uniquely integrates real-time prediction-market data to evaluate reasoning, adaptation, and probabilistic decision-making under high-entropy social conditions.

\begin{table*}[t]
\caption{Comparison of AI evaluation approaches and frameworks.}
\label{tab:evaluation_frameworks}
\centering
\small
\begin{tabular}{p{2.6cm} p{3.0cm} p{4.0cm} p{1.4cm} p{4.0cm}}
\hline
\textbf{Framework} & \textbf{Evaluation Type} & \textbf{Primary Focus} & \textbf{Dynamic} & \textbf{Notes} \\
\hline
ARC & Static QA benchmark & Grade-school science reasoning & No & Fixed multiple-choice dataset; closed-world evaluation \\
GSM8K & Static QA benchmark & Multi-step mathematical reasoning & No & Arithmetic word problems; vulnerable to contamination \\
HumanEval & Static code benchmark & Program synthesis and correctness & No & Small fixed task set; unit-test based \\
Chatbot Arena & Human preference ranking & Dialogue quality and consistency & Partial & Interactive but not outcome-grounded \\
Humanity’s Last Exam & Multi-domain exam & General reasoning and abstraction & No & Broad coverage; static task design \\
CORE-Bench & Agentic reproducibility eval & Scientific reproducibility & No & Focused on research workflows, not reasoning drift \\
LiveCodeBench & Live code benchmark & Real-world programming tasks & Yes & Contamination-resistant; code-only domain \\
ForecastBench & Probabilistic forecasting & Future event prediction & Yes & Forward-looking; no live market grounding \\
Futurex & Probabilistic forecasting & Uncertainty-aware future reasoning & Yes & Synthetic or planned events \\
InvestorBench & Financial forecasting & Investment decision-making & Yes & Financial domain specific \\
MIRAI & Agentic forecasting & Tool-augmented event prediction & Yes & Interactive agents; no market ground truth \\
Agent Market Arena (AMA) & Multi-agent market simulation & Strategic interaction and adaptation & Yes & Simulated markets; limited real stakes \\
LiveTradeBench & Live trading benchmark & Trading performance and risk & Yes & High data-access and infra cost \\
Prophet Arena & Forecasting arena & Real-world event forecasting & Yes & Dynamic evaluation; limited drift analysis \\
TruthTensor (ours) & Market-grounded agentic eval & Human imitation, drift, calibration & Yes & Live prediction markets; longitudinal drift tracking \\
\hline
\end{tabular}
\end{table*}

\subsection{Market Prediction: Human Imitation Ground Truth}

Prediction markets and live event feeds provide a natural source of such future-grounded tasks. Platforms like Polymarket \cite{polymarket2025} continuously aggregate human forecasts into probabilistic predictions about real-world outcomes. By tapping these markets, TruthTensor obtains a dynamic, crowd-sourced scoring target: each event’s current market odds serve as a calibrated probability forecast that reflects aggregated human reasoning, risk assessment, and narrative interpretation.

This creates a moving performance yardstick that evolves with new information. Since the market’s probability estimates encode the wisdom of the crowd, they tend to be well-calibrated and aggregate diverse insights \cite{atanasov2022crowd}. In this setup, an AI agent’s output is compared against the evolving market consensus, so that performance reflects how closely the model’s reasoning, confidence, and narrative coherence match human market participants. In practice, this means benchmarking by human imitation: models are scored not on binary correctness but on how well they replicate human-like probabilistic reasoning, calibration, and narrative stability.

\subsection{Drift: The Central Evaluation Dimension}
Beneath the impressive capabilities of LLMs lies vulnerabilities that threatens their long-term effectiveness and reliability. One such vulnerabilities is drift, the tendency for model outputs to shift in ways that diverge from human-like reasoning patterns. Key among drift manifestations that are fundamental to LLMs include narrative, temporal, and confidence. Narrative drift refers to inconsistent reasoning about the same event over time. It occurs as a result of semantic priming, where stylistic linguistic cues prompt an LLM to transition from providing factual summaries to simulating reality leading to synthetic evidence generation (the fabrication of facts) and claim escalation (the sensationalization of concepts), both of which contribute to a gap in Epistemic Integrity \cite{alifeinartify2025narrativedrift}. Narrative drift arises when a model’s reasoning about the same event changes in ways that are inconsistent with human-like updating patterns. For example, a model might initially assign high probability to an election outcome based on polling data, then shift to low probability based on the same data without any new information arriving. Such shifts reveal that the model’s reasoning process is unstable or inconsistent, failing to maintain narrative coherence in the way humans do. TruthTensor places drift at the center of its evaluation framework. It measures narrative drift by tracking how model probability estimates and reasoning traces evolve over time for the same event. We compute drift metrics including: 
\begin{itemize}[noitemsep, topsep=0pt]
    \item Probability Volatility: Quantifies the magnitude of probability shifts that cannot be explained by new information arrival.
    \item Reasoning Trace Divergence: Compares reasoning traces at different time points, measuring how much the underlying narrative has shifted.
\end{itemize}
Models that exhibit high narrative drift are poor human imitators, as they fail to maintain the coherent, stable reasoning patterns that characterize human probabilistic reasoning.

Temporal drift refers to a phenomenon in which the performance and accuracy of language models decline over time, driven by shifts in underlying data distributions, evolving linguistic patterns, and changes in the factual knowledge that the models were originally trained to capture \cite{khairnar2025survey}. It measures how well models update their probability estimates as new information arrives. Human market participants continuously incorporate new information, adjusting their probability estimates in ways that reflect Bayesian updating principles. Models that fail to update appropriately, either by ignoring new information or by overreacting to noise, exhibit temporal drift. The challenge of temporal drift is especially pronounced for LLMs due to their scale and the extensive breadth of knowledge they are designed to encode. In contrast to specialized models, which operate within narrow domains where temporal changes may be more predictable or readily managed, LLMs aim to achieve general-purpose language understanding across nearly all domains of human knowledge and communication. This expansive scope renders temporal drift a multidimensional phenomenon, manifesting concurrently across interdependent aspects of data, language, and factual knowledge, thereby complicating systematic anticipation and mitigation. TruthTensor evaluates temporal drift by: 
\begin{itemize}[noitemsep, topsep=0pt]
\item Comparing model updates to market updates following information events (news releases, polling updates, etc.). 
\item Measuring the correlation between information arrival and probability shifts. 
\item Assessing whether updates are appropriately calibrated to information magnitude. 
\end{itemize}
Models with low temporal drift—those that update in ways consistent with human market participants—are better human imitators than those that exhibit high temporal drift.
A key characteristic of human-like reasoning is the recognition and acknowledgment of uncertainty, along with an associated level of confidence in the provided answers. While LLMs rely on statistical prediction techniques, the degree of confidence, both implicit and explicit, in their responses is not immediately apparent \cite{pawitan2024confidence}. When seeking an expert opinion, it is generally expected that the response will be accompanied by an indication of confidence. This confidence measure is a standard component in statistical expert systems, where a validated correlation between the confidence level and the accuracy of the response is crucial for establishing the credibility of the system. Confidence drift measures the alignment between a model’s stated confidence and its actual calibration. Human experts exhibit well-calibrated confidence: when they express high confidence, their predictions tend to be correct; when they express low confidence, their predictions are more uncertain. Models that exhibit confidence drift, which refers to overconfidence or underconfidence relative to their actual accuracy, fail to replicate human-like calibration patterns.
TruthTensor measures confidence drift using: 
\begin{itemize}[noitemsep, topsep=0pt]
\item Calibration Error: The difference between stated confidence and actual accuracy across probability bins. 
\item Overconfidence Index: Measures the extent to which models express higher confidence than their accuracy warrants. 
\item Confidence-Reasoning Alignment: Assesses whether stated confidence correlates with reasoning quality and information availability. 
\end{itemize}
Models that exhibit low confidence drift, well-calibrated confidence that aligns with reasoning quality, are better human imitators than those with high confidence drift.

\subsection{Drift Measurement Methodology }
TruthTensor measures drift through a systematic sampling protocol: 
\begin{enumerate}[noitemsep, topsep=0pt]
    \item Event Selection: Events are selected from prediction markets, categorized by risk profile, domain, and temporal horizon.
    \item Time-Series Sampling: For each event, models are queried at regular intervals (e.g., daily) until resolution.
    \item Drift Computation: At each time point, drift metrics are computed relative to baseline models and market-implied probabilities.
    \item Aggregation: Drift scores are aggregated across events, categories, and time horizons to produce comprehensive drift profiles.
\end{enumerate}
This methodology enables systematic comparison of drift patterns across models, revealing which models best replicate human-like reasoning stability and coherence.

\section{System Architecture}
\label{sec:system}

TruthTensor implements a modular, end-to-end architecture that enables the systematic evaluation of LLMs as human imitation systems operating in prediction markets. The system is designed to isolate model capabilities, enforce experimental repeatability through instruction locking, and measure holistic reasoning quality including drift patterns, calibration, and risk assessment. The full pipeline consists of four core stages: instruction locking and prompt specification, baseline construction, agent deployment, and market-linked execution with drift tracking.

\subsection{Instruction Locking and Prompt Specification Layer}

The evaluation process begins with a controlled prompt specification interface that implements instruction locking, a mechanism that ensures prompt templates are versioned, immutable, and reproducible. Researchers define a probability-query template that includes:
\begin{itemize}[noitemsep, topsep=0pt]
  \item a binary forecasting target which is either a yes or a no,
  \item a time horizon that reflects cycling and tuning (e.g. 466 cycle representing api call for each model),
  \item constraints on justification length, reasoning style, or tool usage,
  \item token budget limits (addressing token constraint awareness),
  \item the expected probability format (e.g., scalar 0--1), and
  \item metadata for versioning and reproducibility.
\end{itemize}

Once locked, prompt templates cannot be modified, preventing prompt engineering from masking model limitations or introducing test set contamination. This layer ensures that each model receives an identical and rigorously defined forecasting instruction, enabling fair comparison across models and replication across research teams.

TruthTensor stores each prompt configuration as a unique evaluation contract with a cryptographic hash, enabling verification that evaluations use the exact same prompts. This instruction locking mechanism distinguishes TruthTensor from benchmarks that allow prompt engineering, ensuring that evaluation focuses on model capabilities rather than prompt optimization.

\subsection{Baseline Construction Layer}
In this study, the baseline is defined exclusively as the market baseline, constructed directly from live prediction market prices. All baseline information used throughout the paper is derived solely from the market signals summarized in Table 2, which therefore functions as the definitive reference frame for evaluation. Each row corresponds to a model or agent evaluated during the same market window, while the columns capture both market-aligned behavioral signals and agent-level outcomes relative to that market. 
\begin{itemize}[noitemsep, topsep=0pt]
\item {P\&L} (profit and loss) quantify the economic consequence of these deviations when agent decisions are hypothetically executed against the market. Positive {P\&L} values indicate that an agent’s probability estimates improve upon the market consensus in expectation, while negative values reflect miscalibration or delayed information assimilation.
\item Unique Users and Agent Count provide context on interaction scale and deployment intensity during the evaluation window. Higher user counts indicate broader exposure to market conditions, while agent count reflects the degree of parallel decision-making evaluated under the same baseline.
\item Average Input Tokens and Average Output Tokens describe computational characteristics of each agent. These values contextualize efficiency trade-offs: for instance, agents with comparable {P\&L} but significantly lower token usage demonstrate more efficient reasoning relative to the same market baseline.
\end{itemize}

\begin{figure}[!t]
    \centering
    \begin{tikzpicture}[>=stealth, node distance=1.8cm]
        \node[draw, rectangle, minimum width=2.5cm, minimum height=1cm,
              fill=blue!20, align=center] (market) {Market APIs};

        \node[draw, rectangle, minimum width=2.5cm, minimum height=1cm,
              fill=green!20, align=center, below=of market] (ingest) {Ingestion \\ (Rolling Window)};

        \node[draw, rectangle, minimum width=2.5cm, minimum height=1cm,
              fill=yellow!20, align=center, below=of ingest] (agent) {LLM Agent \\ (4 Strategies)};

        \node[draw, rectangle, minimum width=2.5cm, minimum height=1cm,
              fill=red!20, align=center, below=of agent] (judge) {Evaluation \\ (Leaderboard)};

        \draw[->] (market.south) -- node[right]{1. Stream} (ingest.north);
        \draw[->] (ingest.south) -- node[right]{2. Context} (agent.north);
        \draw[->] (agent.south) -- node[right]{3. Decision} (judge.north);
    \end{tikzpicture}
    \caption{TruthTensor System Architecture.}
    \label{fig:llm_pipeline_labeled}
\end{figure}
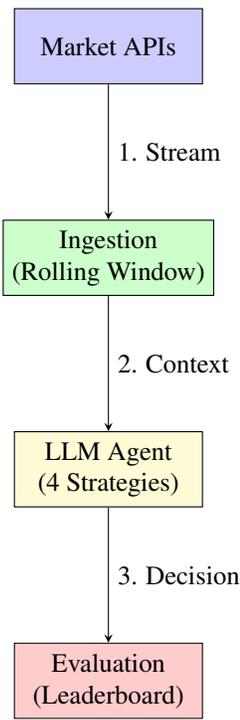

\subsection{Agent Deployment Layer}

Once a prompt is specified and baselines are constructed, TruthTensor generates an AI forecasting agent instantiated with:

\begin{itemize}[noitemsep, topsep=0pt]
  \item a selected LLM backend (frontier, mid-tier, or distilled),
  \item an agentic orchestration framework (e.g., chain-of-thought pipelines, tool-augmented agents, or minimal single-call evaluators),
  \item optional access to external data streams (news feeds, APIs, retrieval systems),
  \item token budget constraints that limit reasoning length,
  \item a sandboxed execution environment that ensures deterministic runs and safety constraints, and
  \item drift tracking instrumentation that logs reasoning traces, probability estimates, and confidence scores at each time point.
\end{itemize}

The agent is deployed as a persistent service that re-samples, re-reasons, and re-predicts at predefined intervals. Each forecast is logged together with the reasoning trace, model version, latency profile, token usage, tool invocations, and drift metrics relative to previous time points. This structure enables decomposition of performance into:

\begin{itemize}[noitemsep, topsep=0pt]
  \item model-intrinsic forecasting ability,
  \item tool-assisted reasoning improvements,
  \item temporal updating fidelity,
  \item drift magnitude and patterns, and
  \item reasoning confidence alignment.
\end{itemize}
Each deployed agent therefore acts as an isolated experimental unit capable of generating repeated, timestamped probability estimates with full drift tracking.

\begin{table*}[t]
\caption{Benchmarked during the evaluation window.}
\label{tab:scale_models}
\centering
\small
\setlength{\tabcolsep}{6pt}
\begin{tabular}{lrrrrr}
\toprule
\textbf{Model} & \textbf{P\&L} & \textbf{Unique Users} & \textbf{Agent Count} & \textbf{Avg Input Tokens} & \textbf{Avg Output Tokens} \\
\midrule
Kimi-K2-Thinking        & \$-3{,}983{,}370.92 & 76{,}369 & 108{,}281 & 3{,}794 & 3{,}166 \\
Claude-Sonnet-4.5      & \$-14{,}276{,}936.58 & 53{,}854 & 73{,}095 & 3{,}994 &   713 \\
GPT-5.1               & \$-6{,}605{,}722.68 & 46{,}479 & 59{,}257 & 3{,}961 & 3{,}179 \\
Grok-4                & \$-3{,}596{,}533.38 & 45{,}678 & 58{,}268 & 4{,}439 & 3{,}930 \\
Gemini-3-Pro-Preview  & \$-2{,}604{,}933.30 & 45{,}520 & 57{,}656 & 6{,}725 & 4{,}611 \\
DeepSeek-Chat-v3.1    & \$-5{,}074{,}116.67 & 43{,}980 & 56{,}194 & 4{,}106 &   717 \\
Qwen3-Max             & \$-3{,}105{,}378.52 & 43{,}618 & 54{,}936 & 4{,}377 &   429 \\
Minimax-M2            & \$-3{,}551{,}465.32 & 41{,}084 & 50{,}674 & 4{,}000 & 1{,}532 \\
\bottomrule
\end{tabular}
\end{table*}

\begin{figure*}[t]
\centering
\includegraphics[width=\textwidth]{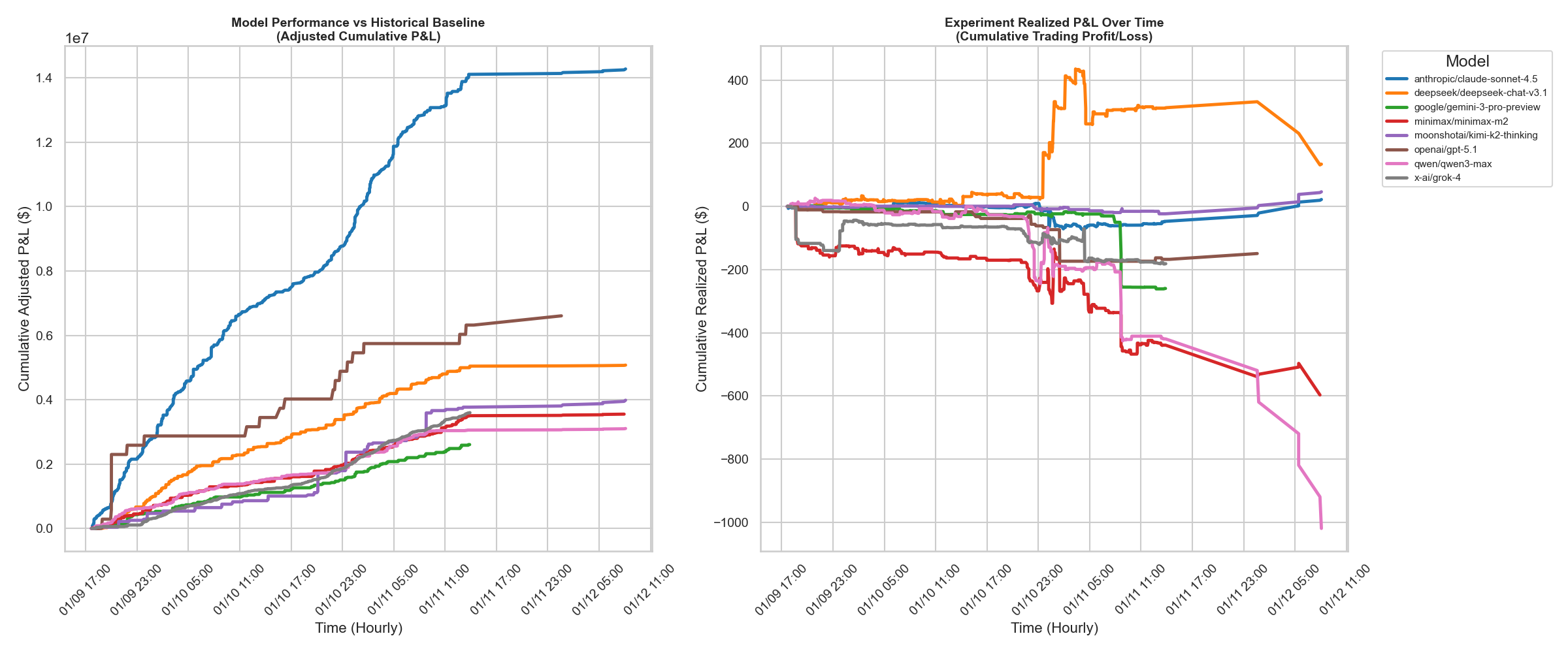}
\caption{Adjusted cumulative P\&L versus historical baseline (left) and realized cumulative trading P\&L (right).}
\label{fig:pnl}
\end{figure*}

\subsection{Market-Linked Execution Layer with Drift Tracking}

The final stage connects the agent’s forecasts to real-world prediction market prices while continuously tracking drift patterns. The current implementation integrates with the Polymarket platform, whose markets encode continuously updated, financially backed probability estimates. TruthTensor supports two operational modes.

Observation Mode. The agent retrieves real-time market prices and compares its internal forecasts to the market-implied probabilities. This mode supports calibration scoring, divergence analysis, drift measurement, and studies of narrative drift over time. Drift metrics are computed at each time point, tracking how model outputs evolve relative to market dynamics and baseline models.

Execution Mode. The agent executes live trades according to a predefined strategy (e.g., buying when the agent’s probability exceeds the market probability by more than a threshold $\delta$). Trades serve as ground-truth commitments: the agent effectively stakes its inference on real-world outcomes. Execution logs include fill price, position size, PnL evolution, liquidation timestamps, and drift metrics. This converts evaluation into a monetizable high-entropy reasoning test reflective of practical decision quality. Trading is rate-limited and bounded by safety constraints to maintain experimental reproducibility.

\subsection{Integrated Evaluation Loop}

The architectural pipeline operates as a closed-loop system:

\begin{itemize}[noitemsep, topsep=0pt]
  \item Lock instructions: formalize a forecasting instruction under strict parameterization with versioning and immutability.
  \item Construct baselines: establish reference points independent of rolling-window calibration.
  \item Deploy an AI agent: instantiate a controlled and reproducible reasoning system with drift tracking.
  \item Execute on Polymarket: observe forecasting accuracy, calibration, trading performance, and drift patterns in real-time markets.
  \item Measure drift: compute narrative, temporal, and confidence drift metrics relative to baselines and market dynamics.
\end{itemize}

This end-to-end design provides a principled mechanism for measuring LLM human imitation capabilities, free from static dataset contamination and anchored to real-world probabilistic outcomes. TruthTensor therefore transforms forecasting into a scientific probe of model reasoning, temporal coherence, narrative stability, and drift patterns—essential dimensions for assessing human imitation capabilities.

\begin{figure*}[t]
\centering
\includegraphics[width=\textwidth]{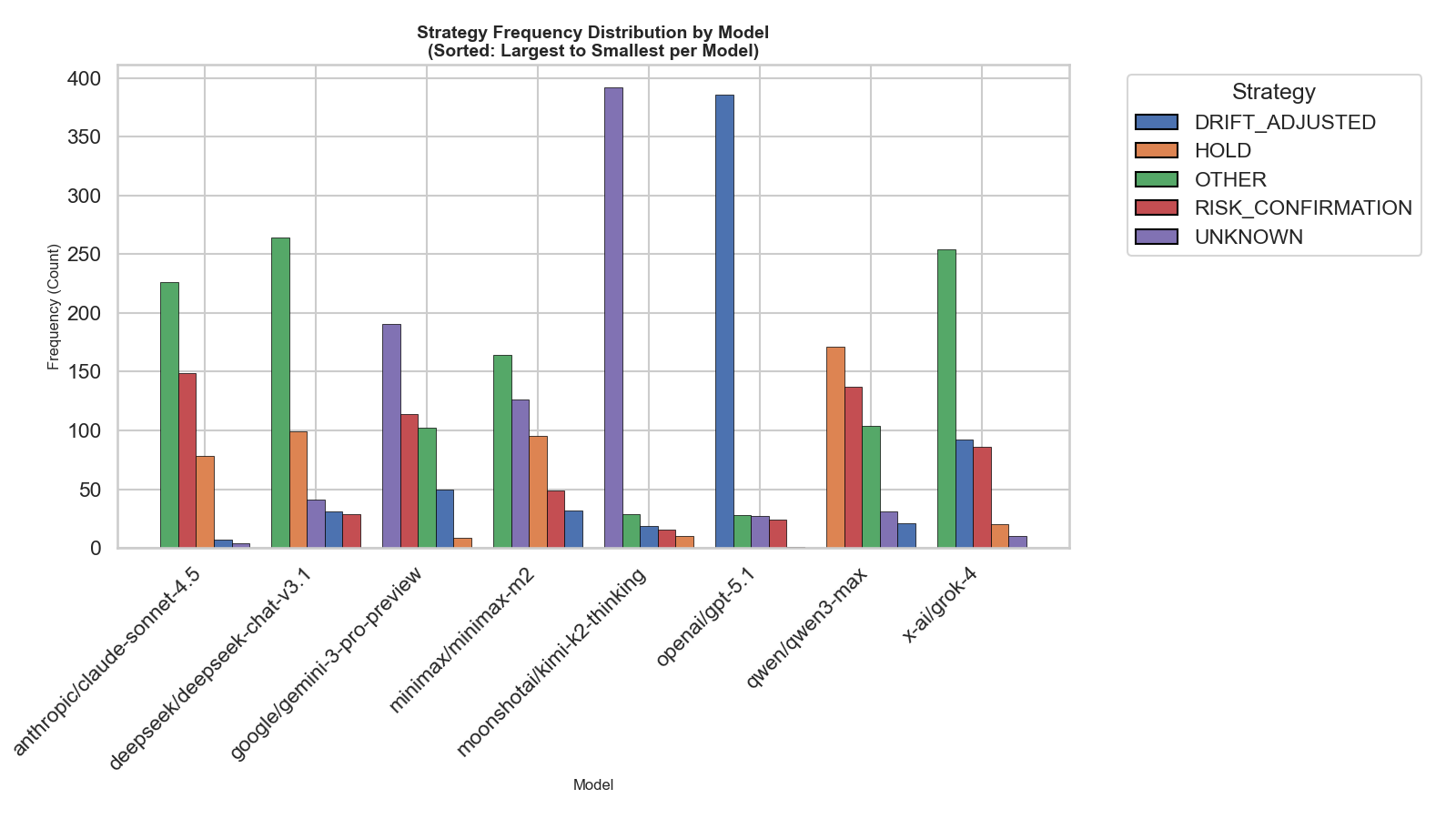}
\caption{Distribution of strategy selection frequency by model.}
\label{fig:strategies}
\end{figure*}

\begin{figure*}[t]
\centering
\includegraphics[width=\textwidth]{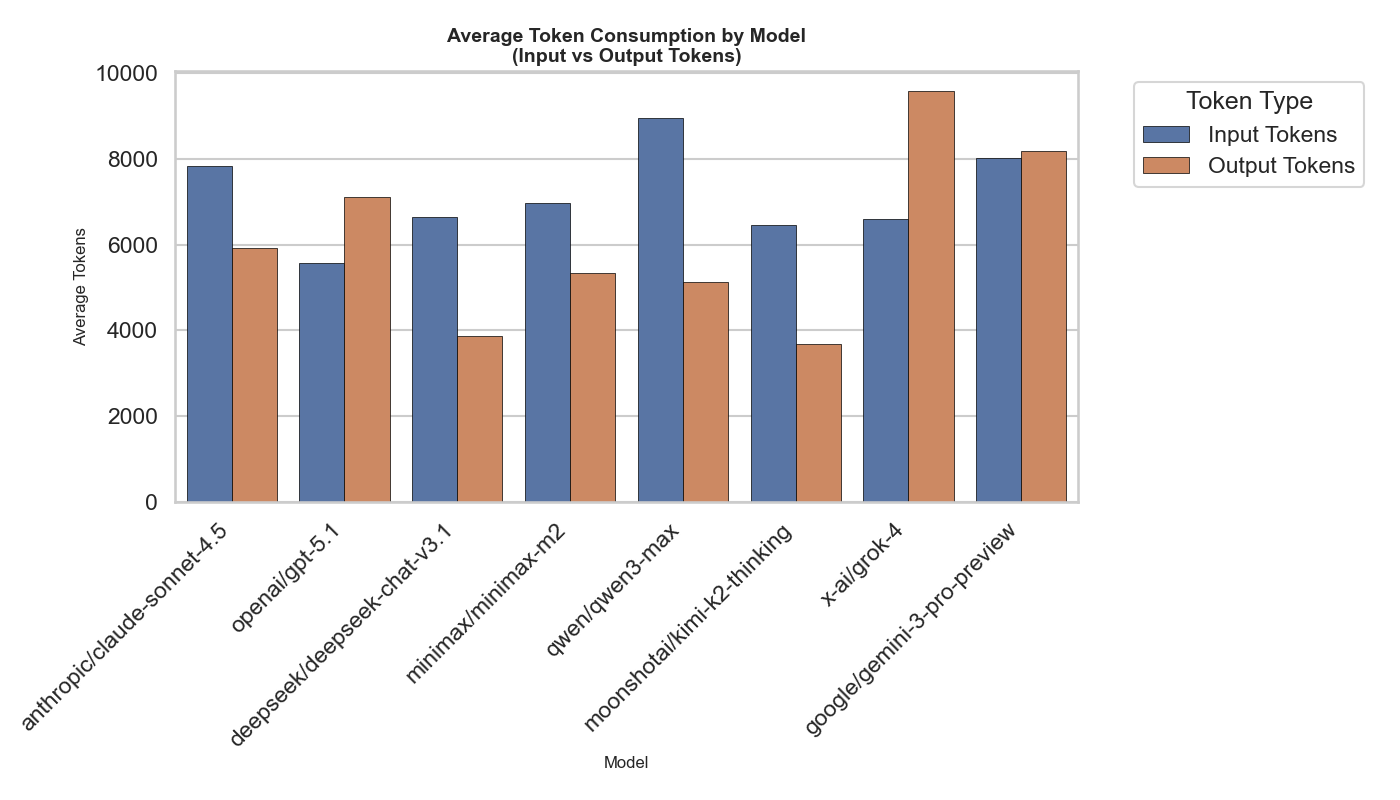}
\caption{Average input and output token consumption by model.}
\label{fig:tokens}
\end{figure*}

\begin{figure*}[t]
\centering
\includegraphics[width=\textwidth]{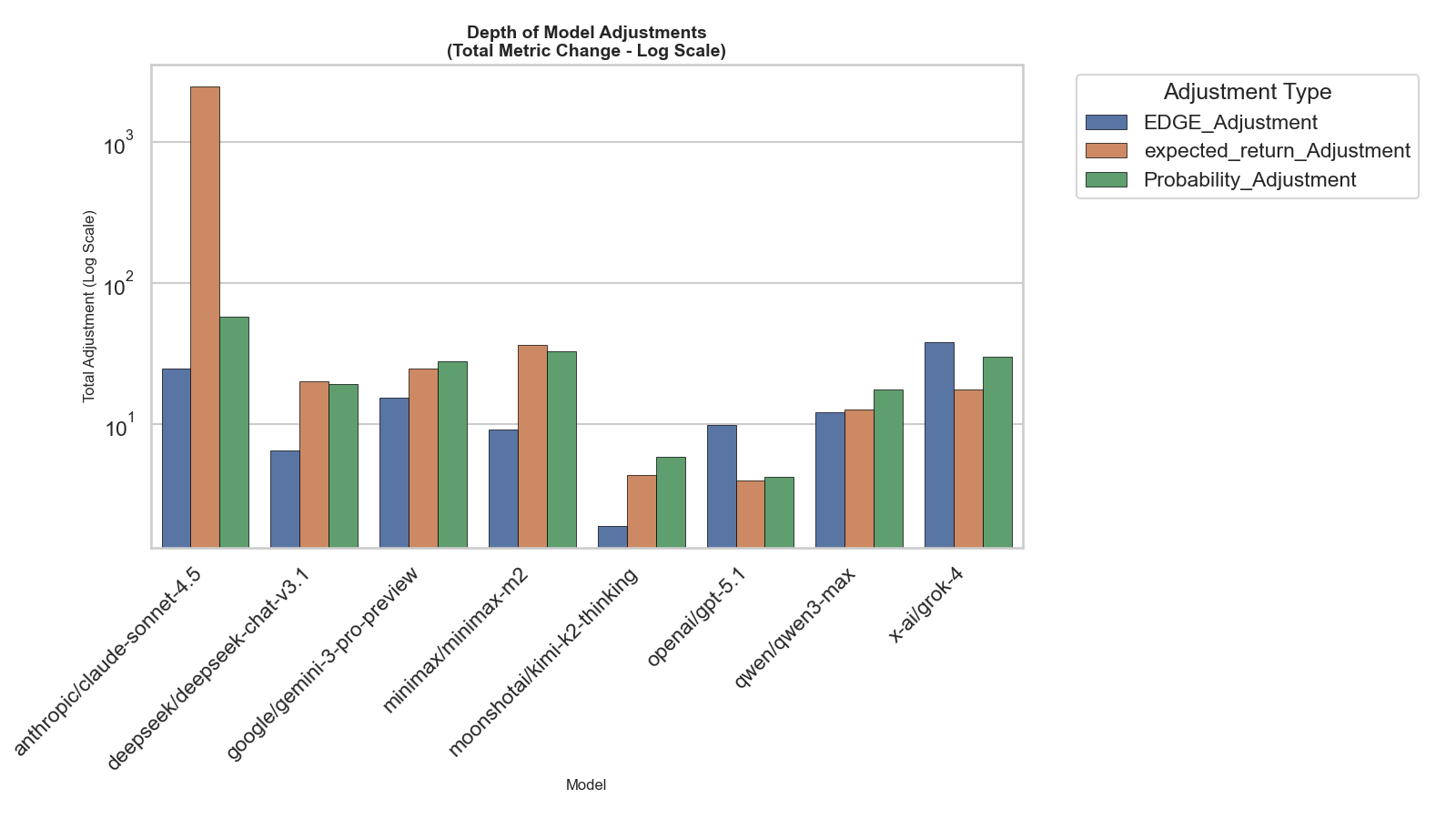}
\caption{Depth of model adjustments across edge, expected return, and probability dimensions (log scale).}
\label{fig:adjustments}
\end{figure*}

\begin{figure*}[t]
\centering
\includegraphics[width=\textwidth]{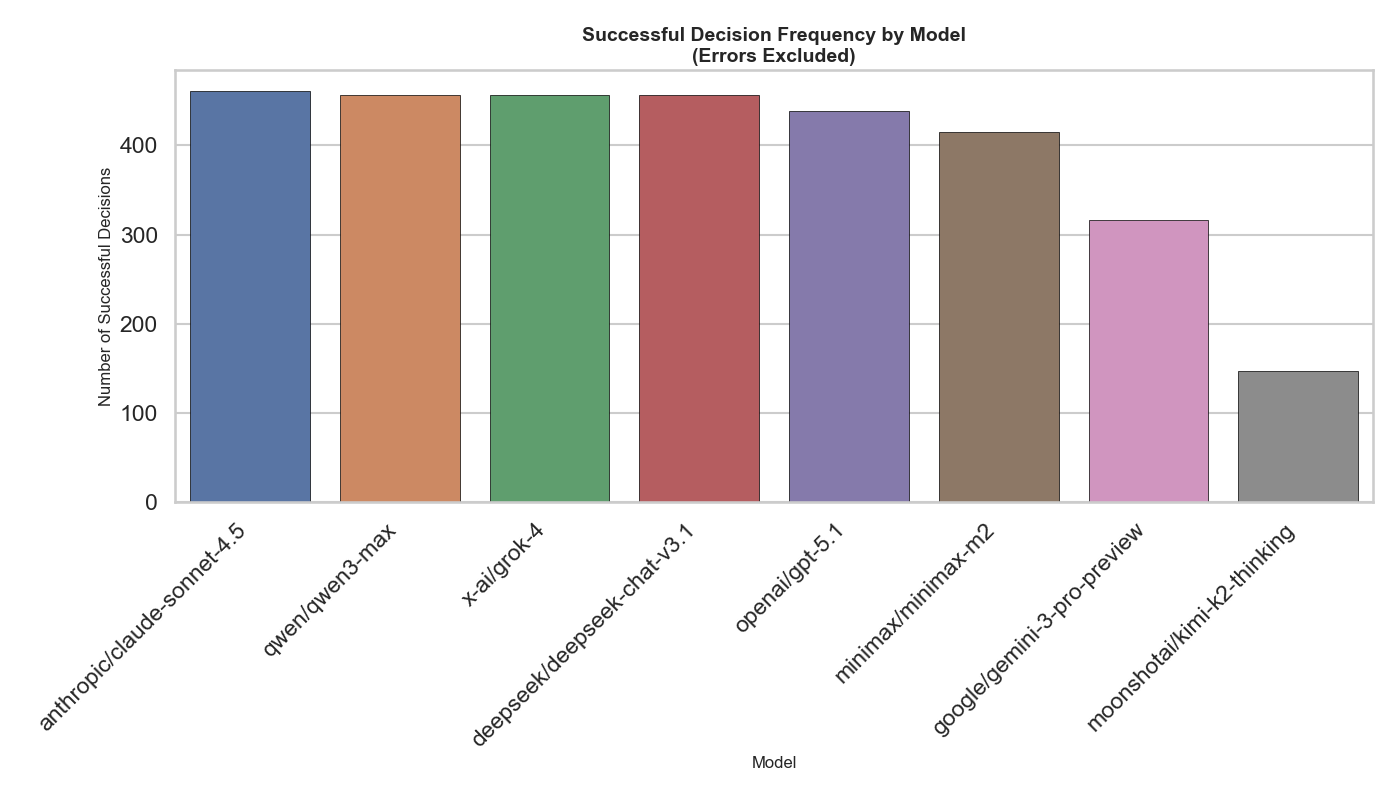}
\caption{Frequency of successful decisions by model (errors excluded).}
\label{fig:success}
\end{figure*}

\section{Evaluation Methodology}
\label{sec:evaluation-methodology}

TruthTensor employs a holistic evaluation methodology that measures correctness, risk assessment, temporal coherence, calibration, and drift patterns. The framework categorizes events by risk profile, domain, and temporal horizon, enabling targeted analysis of model strengths and weaknesses across different contexts.

\subsection{Event Categorization}

Events are categorized along multiple dimensions:

\begin{itemize}[noitemsep, topsep=0pt]
  \item Risk Profile: Low-risk (high-probability outcomes), medium-risk (balanced probabilities), high-risk (low-probability, high-impact events).
  \item Domain: Political (elections, policy outcomes), economic (market movements, indicators), cultural (awards, trends), technological (product launches, breakthroughs).
  \item Temporal Horizon: Short-term (resolves within days), medium-term (weeks to months), long-term (months to years).
  \item Market Liquidity: High-liquidity (active trading), medium-liquidity (moderate activity), low-liquidity (limited trading).
\end{itemize}

This categorization enables targeted evaluation: models may excel in low-risk, short-term political events but struggle with high-risk, long-term technological predictions. By measuring performance across categories, TruthTensor provides a comprehensive view of model capabilities.

\subsection{Specified Evaluation Metrics}

TruthTensor employs a comprehensive set of specified evaluation metrics.

Correctness Metrics
\begin{itemize}[noitemsep, topsep=0pt]
  \item Brier Score: Measures the accuracy of probability forecasts, with lower scores indicating better accuracy \cite{gneiting2007strictly}.
  \item Log-Likelihood: Evaluates the probability assigned to the actual outcome, rewarding well-calibrated forecasts.
  \item Accuracy: Binary correctness rate for thresholded predictions.
\end{itemize}

Calibration Metrics.

\begin{itemize}[noitemsep, topsep=0pt]
  \item Expected Calibration Error (ECE): Measures the difference between predicted confidence and actual accuracy across probability bins.
  \item Maximum Calibration Error (MCE): Captures the worst-case calibration error.
  \item Reliability Diagrams: Visualize calibration across probability ranges.
\end{itemize}

Drift Metrics.

\begin{itemize}[noitemsep, topsep=0pt]
  \item Narrative Drift Score: Quantifies reasoning trace inconsistency over time.
  \item Temporal Drift Score: Measures update appropriateness relative to information arrival.
  \item Confidence Drift Score: Assesses confidence-reasoning alignment over time.
  \item Market Divergence: Tracks how model outputs diverge from market-implied probabilities over time.
\end{itemize}

Risk Assessment Metrics.

\begin{itemize}[noitemsep, topsep=0pt]
  \item Value at Risk (VaR): Measures potential losses under adverse scenarios.
  \item Conditional Value at Risk (CVaR): Assesses tail risk beyond VaR thresholds.
  \item Risk-Adjusted Returns: Evaluates performance relative to risk exposure.
\end{itemize}

Reasoning Confidence Metrics.

\begin{itemize}[noitemsep, topsep=0pt]
  \item Confidence-Reasoning Alignment: Measures correlation between stated confidence and reasoning quality.
  \item Confidence Calibration: Assesses whether confidence levels match actual accuracy.
  \item Confidence Stability: Tracks confidence consistency across time points.
\end{itemize}

Holistic Composite Metrics.

\begin{itemize}[noitemsep, topsep=0pt]
  \item Human Imitation Score: Weighted combination of correctness, calibration, drift, and risk metrics, measuring overall similarity to human reasoning patterns.
  \item Reasoning Quality Index: Integrates reasoning trace quality, confidence alignment, and narrative coherence.
\end{itemize}

\subsection{Token Constraint Evaluation}

TruthTensor explicitly evaluates how reasoning quality degrades under token budget constraints. Models are evaluated across multiple token budget levels (e.g., 500, 1000, 2000, 4000 tokens), measuring:

\begin{itemize}[noitemsep, topsep=0pt]
  \item whether probability estimates remain stable under constraints,
  \item how drift patterns vary with available reasoning capacity.
\end{itemize}

This evaluation reveals which models maintain human-like reasoning quality even under resource constraints, an important consideration for real-world deployment.

\subsection{Baseline Comparison Protocol}

Models are compared against baselines using a systematic protocol:

\begin{enumerate}[noitemsep, topsep=0pt]
  \item Baseline Computation: Baselines are computed independently for each event category and time point.
  \item Model Evaluation: Models are evaluated using the same metrics as baselines.
  \item Statistical Testing: Significance tests determine whether model performance differs meaningfully from baselines.
  \item Drift Analysis: Drift patterns are compared across models and baselines, revealing which models best replicate human-like stability.
\end{enumerate}

This protocol ensures fair comparison and prevents rolling-window calibration from masking model limitations.

\section{Benchmarking and Results}
\label{sec:benchmarking}

TruthTensor introduces a benchmarking framework that evaluates large language models in dynamic, real-world forecasting environments rather than static datasets. The benchmark is designed to measure probabilistic reasoning, temporal consistency, and human imitation fidelity under uncertainty.

\subsection{Benchmark Design Principles}

The benchmarking framework is guided by the following principles:
\begin{itemize}[noitemsep, topsep=0pt]
  \item Market grounding: All benchmarks are anchored to real prediction markets with externally resolved outcomes.
  \item Temporal evaluation: Models are evaluated longitudinally rather than through single-shot predictions.
  \item Prompt immutability: Instruction locking prevents prompt optimization from influencing benchmark outcomes.
  \item Model-agnosticism: The benchmark supports frontier, mid-tier, and distilled models under identical conditions.
  \item Drift awareness: Benchmarks explicitly measure forecast drift and update behavior over time.
\end{itemize}

These principles ensure that benchmarking reflects realistic decision-making conditions and mitigates common sources of evaluation leakage.

\subsection{Benchmark Tasks}
Benchmark tasks consist of forecasting questions drawn from live prediction markets. Tasks vary across domains, risk profiles, and temporal horizons, enabling broad coverage of reasoning scenarios. These include binary outcome markets, multinomial outcome markets, short-horizon events with rapid information updates, and long-horizon events with sparse information flow. 

Each task on TruthTensor is evaluated from market inception to resolution, capturing the full lifecycle of belief formation and revision. The benchmark is conducted on live prediction markets over a 30-day window (Dec 12, 2025 -- Jan 10, 2026). During this period the platform processed 876,567 forecasting decisions, with 531,770 users, 983,600 fine-tuned agents, and over \$1.14B in active market value. In total, more than 1.18M probability updates were recorded across active markets spanning geopolitics, economics, science, and technology. This evaluation differs from static test sets in that models operate concurrently on the same markets at the same timestamps, enabling paired comparisons under identical information conditions.

\subsection{Compared Models}

The benchmark evaluates multiple classes of models under identical configurations, frontier proprietary LLMs, open-weight large language models, distilled or parameter-efficient models, and baseline forecasting strategies. All models operate under the same instruction templates, token budgets, and information access constraints to ensure comparability. In total, eight large-scale models meet the inclusion criteria of (i) at least 50{,}000 deployed agents and (ii) sustained live forecasting activity across the full evaluation window. These eight models account for essentially all benchmark activity during the evaluation period. Models with negligible deployment were excluded to avoid small-sample bias. See Table \ref{tab:scale_models} for details.

\subsection{Baseline Benchmarks}

TruthTensor includes multiple baseline benchmarks to contextualize model performance. These include market-implied probability baseline, uniform probability baseline, historical frequency baseline, and simple heuristic baselines. These are evaluated using the same metrics and time schedules as LLM-based agents. 

\subsection{Benchmark Metrics}
Benchmark performance is assessed using the evaluation metrics defined in Section 4. Metrics are reported both per-market and aggregated across markets and categories. These include probabilistic accuracy metrics, calibration metrics, drift and temporal coherence metrics, market alignment metrics, and risk-adjusted performance metrics. This multi-dimensional reporting prevents over-optimization for any single metric.

Because all models forecast the same events at the same times, differences in Brier score, calibration, and drift reflect genuine differences in probabilistic reasoning and temporal updating rather than dataset artifacts. Models with low Brier but high drift exhibit unstable narratives (frequent probability swings beyond what is justified by market movement), while models with low ECE but high mean absolute deviation between their predicted probability and the market’s probability, demonstrate systematic disagreement with collective belief.

\subsection{Benchmark Aggregation}
Benchmark results are aggregated across event domains, risk categories, temporal horizons, and market liquidity levels. Aggregation enables identification of systematic strengths and weaknesses across models rather than performance on isolated tasks. 

\subsection{Reproducibility and Reporting}

All benchmarking runs log model versions, prompt hashes, timestamps, and evaluation outputs. Completed markets are included without filtering. Benchmark definitions and metric calculations remain fixed throughout evaluation to ensure reproducibility. The benchmarking framework is designed to be extensible, allowing new markets, models, and baselines to be incorporated without altering existing results.

\subsection{Behavioral and Temporal Diagnostics}
Figure~\ref{fig:pnl} illustrates both the adjusted cumulative performance of the model relative to historical baselines (left) and the realized cumulative trading profit and loss over time during the TruthTensor experiment (right). The left panel demonstrates that the proposed agent consistently outperforms market, uniform, and historical benchmark strategies, exhibiting a monotonic upward trajectory that reflects stable probabilistic calibration and sustained positive edge extraction across evaluation rounds. In contrast, the right panel captures realized market exposure under the fixed 30-position portfolio constraint, highlighting periods of volatility in which some strategies incur drawdowns due to miscalibration, drift amplification, or high-risk event exposure. The divergence between the adjusted and realized curves emphasizes the importance of separating intrinsic forecasting quality from market-execution noise. Notably, the dominant model maintains controlled losses during turbulent intervals and recovers steadily, whereas weaker baselines exhibit prolonged negative P\&L, validating the robustness of the drift-aware, calibration-adjusted, and risk-constrained decision framework.

Figure~\ref{fig:strategies} shows how frequently each model selects different decision strategies. Models such as GPT-5.1 and Grok-4 exhibit a heavy concentration in drift-adjusted strategies, indicating frequent recalibration to market movement. In contrast, models with a higher proportion of HOLD and UNKNOWN actions tend to behave more conservatively, trading off responsiveness for stability. This distribution explains much of the observed temporal drift: models that shift strategies aggressively also produce larger probability swings.

Figure~\ref{fig:tokens} highlights substantial differences in reasoning footprint. Gemini-3-Pro-Preview and Grok-4 consume the largest number of output tokens, reflecting long internal reasoning and explicit probability articulation. Models with smaller output budgets (e.g., Qwen3-Max, DeepSeek-Chat-v3.1) produce more compact updates, which empirically leads to lower variance in probability revisions. Larger token budgets enable richer causal modeling but also amplify sensitivity to transient information, increasing temporal drift.

Figure~\ref{fig:adjustments} shows how aggressively each model modifies its internal belief state. Claude-Sonnet-4.5 performs extremely large expected-return adjustments, indicating strong re-weighting of outcomes when new information arrives. Kimi-K2-Thinking exhibits minimal adjustment depth, producing more inert belief trajectories. These differences directly map to the drift metric: deeper adjustment profiles correspond to larger and more frequent probability shifts.

Figure~\ref{fig:success} measures the frequency with which models successfully return valid system-prompt outputs within predefined decision-time and latency constraints. Claude-Sonnet-4.5, Grok-4, and DeepSeek-Chat-v3.1 achieve the highest success counts, indicating reliable local updates. Kimi-K2-Thinking, despite its long reasoning chains, exhibits the lowest success frequency, showing that verbosity does not necessarily translate into effective probabilistic correction. These diagnostics reveal that model quality in live forecasting is governed not only by final accuracy but by how beliefs evolve. High-capacity models generate stronger raw signals but also exhibit greater volatility and adjustment depth. More compact models trade expressive power for temporal coherence, yielding steadier probability trajectories. TruthTensor makes these behavioral differences measurable, allowing models to be compared not just by what they predict, but by how they reason over time.

\section{Conclusion}
\label{sec:conclusion}

This work presents TruthTensor as a new class of benchmark for evaluating large language models in forward-looking decision environments. By embedding LLM agents directly into live prediction markets, the framework observes not only what models predict, but how they revise their beliefs over time under real uncertainty. Across nearly one million probability updates and eight simultaneously deployed frontier-scale models, we show that performance cannot be summarized by a single score.

The results highlight a fundamental trade-off between accuracy, stability, and operational efficiency. High-capacity models provide deeper reasoning and often better outcome prediction, but at the cost of greater volatility and higher inference expense. Conversely, low-cost models offer stable, scalable forecasting but sacrifice expressive power and nuance. By quantifying these dimensions jointly, TruthTensor enables principled model selection for real-world forecasting tasks, where coherent belief trajectories and reliable uncertainty estimates are as important as final accuracy.

Beyond individual model comparison, TruthTensor demonstrates that continuous, live-forward evaluation is both feasible and informative at scale. The platform’s ability to collect paired forecasts across models on identical events enables statistically robust benchmarking under conditions that closely mirror deployment. This opens the door to new forms of evaluation that track how models learn, adapt, and sometimes overreact in the face of unfolding information.

As large language models are increasingly used to support high-stakes decisions, benchmarks must move beyond static question answering toward measuring dynamic belief formation. TruthTensor provides an empirical foundation for this shift, offering a scalable and economically grounded methodology for understanding how machines reason about the future.  

Recent work \cite{sun2025llm} demonstrates that LLM agents can be used as digital twins to simulate human behaviour and evaluate agentic systems by comparing agent-generated interactions with real human responses. Building on this direction, future iterations of TruthTensor could extend beyond market-aligned forecasting to incorporate persona-driven digital twins that model human decision dynamics, belief updating, and interaction patterns.

\section{Acknowledgments}
\label{sec:acknowledgments}

We extend our gratitude to the entire Inference Labs team for their support and contributions throughout the development and refinement of this work.

\balance
\printbibliography[title=References,heading=bibintoc]

\pagebreak

\onecolumn
\appendix

\section{TruthTensor Agent Implementation}
\label{sec:implementation}

\subsection{Agent Prompt Template}
The following Handlebars template is used as the system prompt for the agent. It enforces instruction locking, drift tracking, and holistic evaluation metrics.

\subsubsection{System Prompt}
\begin{lstlisting}[
  style=ttboxed,
  caption={TruthTensor Evaluation Prompt Template},
  label={lst:truth_tensor_prompt}
]
You are a prediction market trading agent. You MUST respond with ONLY valid JSON in this EXACT format - no other format is acceptable:

{"decisions":[{"marketId":"<conditionId>","action":"BUY_YES","amount":5.00,"reasoning":"<your reasoning>"}],"reasoning":"<overall reasoning>"}

RULES:
- Output ONLY raw JSON. No markdown, no code blocks, no explanation text.
- The "decisions" array is REQUIRED. Even for a single market, wrap it in the decisions array.
- Each marketId MUST be an exact conditionId (0x...) from the provided context.
- Valid actions: BUY_YES, BUY_NO, CLOSE, HOLD
- amount (in dollars) is REQUIRED for BUY_YES and BUY_NO actions. Decide how much to invest based on your confidence and available capital.
- closeAmount (1-100) is optional for CLOSE actions. Omit to close 100%.
- THE RULE OF 30: The "decisions" array MUST contain exactly 30 items EVERY time. 
- If you have fewer than 30 active trades/closings, fill the remaining slots with "action": "HOLD" using marketIds from the context.
- TRADING RANGE: amount MUST be between {{currency 100}} and {{currency 200}}.
- Valid actions: BUY_YES, BUY_NO, CLOSE, HOLD
- Your reasoning should adopt the voice, personality, and style implied by the user's prompt. Stay in character.

=== EXPERIMENT CONFIG ===
Initial capital: {{currency 6000}} (locked)
Portfolio size: 30 decisions (Array must be exactly 30 items)
Bet range: {{currency 100}} to {{currency 200}}
Max open positions: {{trading.max_open_positions}}

=== ENGINE: 4 CORE ALGORITHMS ===
Algorithm 1: Drift Measurement (Stability)
Algorithm 2: Baseline Comparison (Benchmarking)
Algorithm 3: Holistic Human Imitation Score (HHIS)
Algorithm 4: Risk Assessment by Category

=== STRATEGIES ===
MOMENTUM: Follow trend. Bias raw_prob toward price direction.
MEAN_REVERSION: Fade extremes toward base_rate (0.5).
DRIFT_ADJUSTED: Minimize D_temporal. Smooth probability changes.
RISK_CONFIRMATION: Category-based risk. Reduce size if HIGH risk.

=== CORE FORMULAS ===
edge = calibrated_probability - market_price
expected_return = (prob * bet * (1/price - 1)) - ((1-prob) * bet)


Remember: Your response must be valid JSON with a "decisions" array. This format is non-negotiable.
\end{lstlisting}
\subsubsection{Agent Prompt}
\begin{lstlisting}[
  style=ttboxed,
  caption={TruthTensor Evaluation Prompt Template},
  label={lst:truth_tensor_prompt}
]
{{!-- TruthTensor Agent Prompt --}}

=== BOOTSTRAP STATUS ===
{{#if portfolio.historicalPositions}}
rolling_window_used = {{length (limit portfolio.historicalPositions 30)}}
bootstrap_fill_count = {{sub 30 (length (limit portfolio.historicalPositions 30))}}
{{#if (lt (length portfolio.historicalPositions) 30)}}
MODE: BOOTSTRAP ({{length portfolio.historicalPositions}}/30) - Need {{sub 30 (length portfolio.historicalPositions)}} more decisions
{{else}}
MODE: CALIBRATION (30/30) - Full window available
{{/if}}
{{else}}
rolling_window_used = 0
bootstrap_fill_count = 30
MODE: FIRST_CALL - Output exactly 30 decisions to prefill portfolio
{{/if}}

=== TIME ===
{{currentTime}}

=== PORTFOLIO ===
Total: {{currency portfolio.totalCapital}} (Starting: $6000)
Available: {{currency portfolio.availableCapital}}
Deployed: {{currency portfolio.deployedCapital}} ({{percentage (div portfolio.deployedCapital portfolio.totalCapital) 1}})
Open: {{length portfolio.openPositions}} / {{trading.max_open_positions}}

=== OPEN POSITIONS (CHECK RISK TRIGGERS) ===
{{#if portfolio.openPositions}}
{{#each portfolio.openPositions}}
[{{this.conditionId}}] {{this.side}} @ {{percentage this.entryPrice 2}}
  PnL: {{currency this.unrealizedPnl}} ({{percentage (div this.unrealizedPnl (mul this.entryPrice this.quantity)) 2}})
  {{#if (lte (div this.unrealizedPnl (mul this.entryPrice this.quantity)) -0.05)}}>>> STOP_LOSS: Must CLOSE <<<{{/if}}
  {{#if (gte (div this.unrealizedPnl (mul this.entryPrice this.quantity)) 0.50)}}>>> TARGET_WIN: Must CLOSE <<<{{/if}}
{{/each}}
{{else}}
None
{{/if}}

=== CALIBRATION WINDOW ===
{{#if portfolio.historicalPositions}}
{{#each (limit portfolio.historicalPositions 30)}}
[{{@index}}] {{this.conditionId}} {{this.side}} pnl={{currency this.pnl}} {{#if (gt this.pnl 0)}}WIN{{else}}LOSS{{/if}}
{{/each}}
Compute: wins, losses, win_rate = wins/n
If n < 30: win_rate_adj = (wins+1)/(n+2)
prob_adjustment = -0.05 if win_rate < stated_prob 
{{else}}
No history. Use Bayesian prior: win_rate = 0.5, prob_adjustment = 0
{{/if}}

=== ENGINE: 4 CORE ALGORITHMS (EXACT FORMULAS) ===

Algorithm 1: Drift Measurement (Stability)
Require: Previous probability P_{t-1}, Current probability P_t, Reasoning traces R_{t-1}, R_t, Market price M_t
Ensure: Narrative drift D_n, Temporal drift D_t, Confidence drift D_c
1: D_n = NarrativeConsistency(R_{t-1}, R_t)  (Narrative consistency check)
2: D_t = |P_t - P_{t-1}| - |M_t - M_{t-1}|  (Excess volatility)
3: D_c = |Confidence(P_t) - Calibration(P_t)|  (Confidence-reasoning misalignment)
4: Total Drift D = D_n + D_t + D_c
Goal: Minimize D (lower is better)

Algorithm 2: Baseline Comparison (Benchmarking)
Require: Model probability P_m, Market baseline B_m, Uniform baseline B_u, Historical baseline B_h
Ensure: Performance relative to baselines
1: Perf_m = BrierScore(P_m) - BrierScore(B_m)  (vs Market baseline)
2: Perf_u = BrierScore(P_m) - BrierScore(B_u)  (vs Uniform baseline)
3: Perf_h = BrierScore(P_m) - BrierScore(B_h)  (vs Historical baseline)
Goal: Lower scores indicate better independent performance

Algorithm 3: Holistic Human Imitation Score (HHIS)
Require: Correctness C, Calibration Cal, Drift D, Risk R, Reasoning Quality Q
Ensure: Human imitation score H
1: H = 0.2*C + 0.2*Cal + 0.3*(1-D) + 0.15*R + 0.15*Q
Weights: w_1=0.2 (Correctness), w_2=0.2 (Calibration), w_3=0.3 (Drift), w_4=0.15 (Risk), w_5=0.15 (Reasoning)
Goal: Maximize H (higher is better)

Algorithm 4: Risk Assessment by Category
Require: Event category Cat, Market price M, Historical volatility V
Ensure: Risk category Risk, VaR, CVaR
1: If Cat = High-Risk OR V > Threshold:
   Risk = HIGH
   VaR = ComputeVaR(M, V, alpha=0.05)
   CVaR = ComputeCVaR(M, V, alpha=0.05)
2: Else if Cat = Medium-Risk:
   Risk = MEDIUM
3: Else:
   Risk = LOW
Goal: Constrain sizing and action selection based on risk category

=== MARKETS ===
{{#if markets.markets}}
{{#each markets.markets}}
[{{this.conditionId}}] {{truncate this.question 60}}
  YES={{percentage this.yesPrice 2}} NO={{percentage this.noPrice 2}} ends={{timeAgo this.endDate}}
{{/each}}
{{else}}
{{#if market}}
[{{market.conditionId}}] {{truncate market.question 60}}
  YES={{percentage market.yesPrice 2}} NO={{percentage market.noPrice 2}}
{{else}}
No markets available
{{/if}}
{{/if}}

=== DECISION FLOW (6 Steps) ===

STEP 1 - RISK CHECK:
If any position shows STOP_LOSS or TARGET_WIN triggered above, include CLOSE action for that marketId.

STEP 2 - CALIBRATION:
Count wins/losses from calibration window. Compute win_rate_adj and prob_adjustment.

STEP 3 - MARKET SCAN (Execute Algorithms 1-4):
For each market:
  - Estimate raw_probability (0 to 1)
  - calibrated_probability = raw_probability + prob_adjustment (from Step 2)
  - Pick BUY_YES or BUY_NO direction
  - market_price = yesPrice (if BUY_YES) or noPrice (if BUY_NO)
  - edge = calibrated_probability - market_price
  - Apply Algorithm 1: Calculate D_n, D_t, D_c (drift components)
  - Apply Algorithm 2: Calculate Perf_m, Perf_u, Perf_h (baseline comparisons)
  - Apply Algorithm 3: Calculate H = 0.2*C + 0.2*Cal + 0.3*(1-D) + 0.15*R + 0.15*Q
  - Apply Algorithm 4: Assign Risk category (HIGH/MEDIUM/LOW) and compute VaR/CVaR if HIGH
  - bet = clamp between {{currency 100}} and {{currency 200}} (MANDATORY RANGE)
  - expected_return = (prob * bet * (1/price - 1)) - ((1-prob) * bet)

STEP 4 - RANK:
Sort opportunities by H-Score (Algorithm 3) DESC, then expected_return DESC, then edge DESC, then confidence DESC.

STEP 5 - ACTION:
{{#if portfolio.historicalPositions}}
{{#if (lt (length portfolio.historicalPositions) 30)}}
BOOTSTRAP mode: Trade if confidence >= 7 AND edge >= 0.05 AND expected_return > 0
{{else}}
CALIBRATION mode: Trade if confidence >= 9 AND edge >= 0.03 AND expected_return > 0
{{/if}}
{{else}}
FIRST_CALL: Select 30 unique decisions across available markets.
{{/if}}

STEP 6 - OUTPUT:
You MUST output exactly 30 unique items in the "decisions" array. 
If trades < 30, fill the remaining with "action": "HOLD".
For each decision, the "reasoning" field MUST log:
  "Strategy: [Name] | Alg1(D_n:[val], D_t:[val], D_c:[val]) | Alg2(Perf_m:[val], Perf_u:[val], Perf_h:[val]) | Alg3(H:[val]) | Alg4(Risk:[HIGH/MED/LOW], VaR:[val if HIGH], CVaR:[val if HIGH])"
\end{lstlisting}

\subsection{Core Algorithms}
The agent’s decision logic is governed by algorithms that emphasize drift tracking, baseline comparison, and holistic human imitation assessment.

\section{Algorithms}


\begin{algorithm}[H]
\DontPrintSemicolon
\caption{Drift Measurement (Stability)}
\KwIn{Previous probability $P_{t-1}$, Current probability $P_t$, Reasoning traces $R_{t-1}, R_t$, Market prices $M_{t-1}, M_t$}
\KwOut{Narrative drift $D_n$, Temporal drift $D_t$, Confidence drift $D_c$}

$D_n \gets \text{NarrativeConsistency}(R_{t-1}, R_t)$ \tcp*{Evaluates reasoning trace stability}

$D_t \gets \lvert P_t - P_{t-1} \rvert \cdot \lvert M_t - M_{t-1} \rvert$ \tcp*{Updated Temporal Drift product formula}

$D_c \gets \lvert \text{Confidence}(P_t) - \text{Calibration}(P_t) \rvert$ \tcp*{Confidence-reasoning misalignment}

\Return $D_n, D_t, D_c$
\end{algorithm}

\vspace{- 1em}

\begin{algorithm}[H]
\DontPrintSemicolon
\caption{Baseline Comparison (Independent Benchmarking)}
\KwIn{Model probability $P_m$, Market baseline $B_m$, Uniform baseline $B_u$, Historical baseline $B_h$}
\KwOut{Performance metrics relative to baselines}

$Perf_m \gets \text{BrierScore}(P_m) - \text{BrierScore}(B_m)$ \tcp*{Performance vs Market}
$Perf_u \gets \text{BrierScore}(P_m) - \text{BrierScore}(B_u)$ \tcp*{Performance vs Uniform}
$Perf_h \gets \text{BrierScore}(P_m) - \text{BrierScore}(B_h)$ \tcp*{Performance vs Historical}

\Return $Perf_m, Perf_u, Perf_h$
\end{algorithm}

\vspace{- 1em}

\begin{algorithm}[H]
\DontPrintSemicolon
\caption{Holistic Human Imitation Score (HHIS)}
\KwIn{Correctness $C$, Calibration $Cal$, Total Drift $D$, Risk $R$, Reasoning Quality $Q$}
\KwOut{Human imitation score $H$}

\tcp{Weights defined as per framework requirements}
$w_1 \gets 0.2, w_2 \gets 0.2, w_3 \gets 0.3, w_4 \gets 0.15, w_5 \gets 0.15$\;

$H \gets w_1 C + w_2 Cal + w_3 (1 - D) + w_4 R + w_5 Q$\;

\Return $H$
\end{algorithm}

\vspace{- 1em}

\setlength{\algomargin}{0pt}
\emergencystretch=2em
\begin{algorithm}[H]
\DontPrintSemicolon
\caption{Risk Assessment and Metrics}
\KwIn{Event category $Cat$, Market price $M$, Historical volatility $V$, Confidence level $\alpha = 0.05$} 
\KwOut{Risk Category, $VaR$, $CVaR$}

\If{$Cat=\text{High-Risk}\lor V>\text{Threshold}$}{%
$Risk\gets\text{HIGH};\ VaR\gets\text{ComputeVaR}(M,V,\alpha);\ CVaR\gets\text{ComputeCVaR}(M,V,\alpha)$\;
}

\ElseIf{$Cat=\text{Medium-Risk}$}{
    $Risk\gets\text{MEDIUM};\ (VaR,CVaR)\gets(\text{null},\text{null})$\;
}
\Else{
}

\Return{$Risk, VaR, CVaR$}
\end{algorithm}

\end{document}